\documentclass[letterpaper, 10 pt, conference]{ieeeconf}  

\IEEEoverridecommandlockouts                              %

\usepackage[utf8]{inputenc}

\title{\LARGE \bf
Human-Robot Shared Control for Surgical Robot Based on Context-Aware Sim-to-Real Adaptation
}

\author{Dandan Zhang${^*}$, Zicong Wu${^*}$, Junhong Chen${^*}$, Ruiqi Zhu,  Adnan Munawar, Bo Xiao, Yuan Guan, \\ Hang Su,~\IEEEmembership{Member, IEEE},  Wuzhou Hong, Yao Guo,~\IEEEmembership{Member, IEEE},   Gregory S. Fischer,~\IEEEmembership{Member, IEEE}, \\Benny Lo,~\IEEEmembership{Senior Member, IEEE}, Guang-Zhong Yang,~\IEEEmembership{Fellow, IEEE} 
\thanks{
$^{*}$These authors contributed to this paper equally. 
\newline
D. Zhang is with the Department of Engineering Mathematics, University of Bristol, United Kingdom. \textit{corresponding email}: ye21623@bristol.ac.uk
\newline
D. Zhang, Y. Guan are affiliated
with the Bristol Robotics Lab, United Kingdom.
Z. Wu, J. Chen, R. Zhu, B. Xiao are with the Hamlyn Centre for Robotic Surgery, Imperial College London, United Kingdom. 
H. Su is with the Department of Electronics, Polytechnic University of Milan, Italy.
W. Hong, Y. Guo,   G.-Z. Yang are  with the Institute of Medical Robotics, Shanghai Jiao Tong University, China. 
A. Munawar and G.S. Fischer are with the Worcester Polytechnic Institute, USA. 
}}

\usepackage{graphicx} 
\graphicspath{ {Figures/} }

\usepackage[caption=false,font=footnotesize]{subfig}
 \usepackage{algorithm,algorithmic}
\usepackage{cite}
\usepackage[font=footnotesize]{subfig}
\usepackage{multirow}
\usepackage{ifpdf}
\usepackage{epstopdf}
\usepackage[cmex10]{amsmath}
\usepackage{algorithmic}
\usepackage{array}
\usepackage{mdwmath}
\usepackage{mdwtab}
\usepackage{eqparbox}
\usepackage{verbatim}

\usepackage{stfloats}
\usepackage[font=footnotesize]{subfig}

\usepackage{comment}
\usepackage{amsmath,bm}
\usepackage{amssymb}
\usepackage{color}
\usepackage{subfig}

\begin{document}

\maketitle

\begin{abstract}
Human-robot shared control, which integrates the advantages of both humans and robots,  is an effective approach to facilitate efficient surgical operation.
Learning from demonstration (LfD) techniques can be used to automate some of the surgical subtasks for the construction of the shared control framework. However, a sufficient amount of data is required for the robot to learn the manoeuvres. Using a surgical simulator to collect data is a less resource-demanding approach. With sim-to-real adaptation, the manoeuvres learned from a simulator can be transferred to a physical robot. To this end, we propose a  sim-to-real adaptation method to construct a human-robot shared control framework for robotic surgery.  

In this paper, a desired trajectory is generated from a simulator using LfD method, while dynamic motion primitive (DMP) based method is used to transfer the desired trajectory from the simulator to the physical robotic platform. Moreover, a role adaptation mechanism is developed such that the robot can adjust its role according to the surgical operation contexts predicted by a neural network model. 
The effectiveness of the proposed framework is validated on the da Vinci Research Kit (dVRK). Results of the user studies indicated that with the adaptive human-robot shared control framework, the path length of the remote controller, the total clutching number and the task completion time can be reduced significantly. The proposed method outperformed the traditional manual control via teleoperation.
\end{abstract}

\maketitle

\section{Introduction}

Surgical robots have been widely adopted in clinical practice, as they can provide improved precision, easing the surgical operations and leading to better  clinical outcomes \cite{yang2018grand,gao2021progress}. 
Most of the existing robotic platforms for surgery  are developed based on teleoperation \cite{zhang2019design,Zhang2020Hamlyn,zhang2019handheld}. However, the current trend for the development of surgical robots is towards a higher level of autonomy, with Artificial Intelligence (AI) incorporated into   robotic systems\cite{peters2018review,yang2017medical}.    For safety consideration, a practical approach is to share autonomy between human operators and robots \cite{payne2021shared}.   The human operator has better cognitive abilities while the robot can ensure a higher level of precision for manipulation and execute some repetitive motions to reduce the operator's burden. Therefore, the human operator and the robot can be complementary to each other, which can lead to enhanced efficiency  during surgical operations and bring more clinical outcomes \cite{li2013building,amirshirzad2019human}. 

Learning from demonstration (LfD) is the paradigm where robots obtain new skills by imitating demonstrations provided by humans \cite{zhang2021explainable,chen2020supervised, ravichandar2020recent}.
To automate some of the surgical sub-tasks for human-robot shared control, LfD approaches can be used. Continuous Hidden Markov Model (CHMM) has been used for
the construction of a cooperative control framework for haptic guidance in robotic surgery \cite{power2015cooperative}. To realize autonomous tissue manipulation, LfD has been used to initialize the dynamics of the tissue and ensure better controller
performance
\cite{shin2019autonomous}. An iterative technique has been explored to learn a desired trajectory for a suture knot tying task \cite{van2010superhuman}. A trajectory transfer algorithm through non-rigid registration has been developed to extract trajectories from demonstration data and adapt them to  new environments for suturing tasks \cite{schulman2013case}. However, the methods mentioned above require demonstration data obtained from the physical robotic systems for model training.   

Instead of using a physical robotic system, the use of a surgical simulator is more efficient and cost-effective in collecting data for LfD. For example, it avoids the tear and wear issues of the surgical tools. However, due to the domain gap between the simulation and the physical environment, the model obtained from a simulator for human-robot shared control cannot be applied to a physical robotic system directly.
Therefore, it is significant to investigate sim-to-real adaptation for the construction of a human-robot shared control framework.


Dynamic Motion Primitives (DMPs) can model a complex dynamic system via a set of nonlinear differential equations. DMP has been used for robot to learn desired motions faithfully from a demonstration \cite{li2020skill}.  Moreover, it can be used to generalise the learned trajectory towards a new goal. Therefore, we incorporate DMP with trajectory spatial transformation and hand-eye coordination to realize the sim-to-real transfer of  the human-robot shared control framework.
 To ensure that the  relative roles of the human operator and the surgical robot can be adjustable,  a role adaptation mechanism should be developed and integrated into the shared control framework.  To this end, machine learning based context-awareness will be investigated in this paper, which can support the implementation of the role adaptation mechanism \cite{wang2021real,zhang2021surgical}.  With accurate recognition of the contexts, the whole surgical procedure can be segmented into different sub-tasks automatically, while adaptive human-robot shared control can be achieved based on the characteristics of the specific surgical sub-tasks.
 


The main contribution of this paper is to develop a context-aware sim-to-real adaptation method for the construction of a human-robot shared control framework, which can be used to enhance the efficiency of surgical operation. 


\section{Methodology}\label{SemiAutonomous}

\subsection{Overview}
\begin{figure*}[htpb!] 
	\centering 
	\includegraphics[width = 1\hsize]{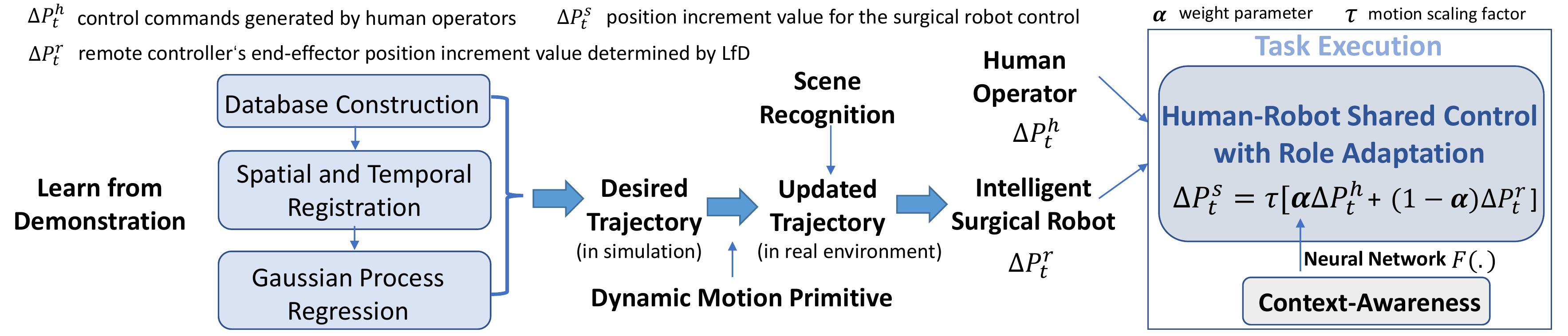} 
	\caption{The framework of the proposed context-aware sim-to-real adaptation method in this paper.  } 
	\label{Fig.3}
\end{figure*}


The framework of the proposed method is illustrated in Fig.~\ref{Fig.3}.
To develop the human-robot shared control framework, the first step is to construct a database, which can be used for  LfD and model training  for context-awareness. In the second step, spatial and temporal registration is applied to the recorded trajectories, while Gaussian Process Regression (GPR) is performed on the registered data to generate the desired trajectory for task execution. Following that, the desired trajectory obtained via GPR is transferred to a physical robot based on DMP with the support of trajectory spatial transformation and hand-eye coordination in the third step. The desired trajectory can be adapted to different initial configurations and goals using DMP, while the vision-based scene recognition method can be used to identify the goal positions for trajectory adaptation.
Finally, to construct the shared control framework for task execution, the surgical robot control commands are determined by a convex combination of the commands generated by the intelligent robot and the human control commands generated via the remote controllers. The control commands generated by the intelligent surgical robot are determined based on the desired trajectory obtained via GPR and DMP.


\subsection{Database Construction}

\begin{figure*}[htpb!] 
	\centering 
	\includegraphics[width = 1\hsize]{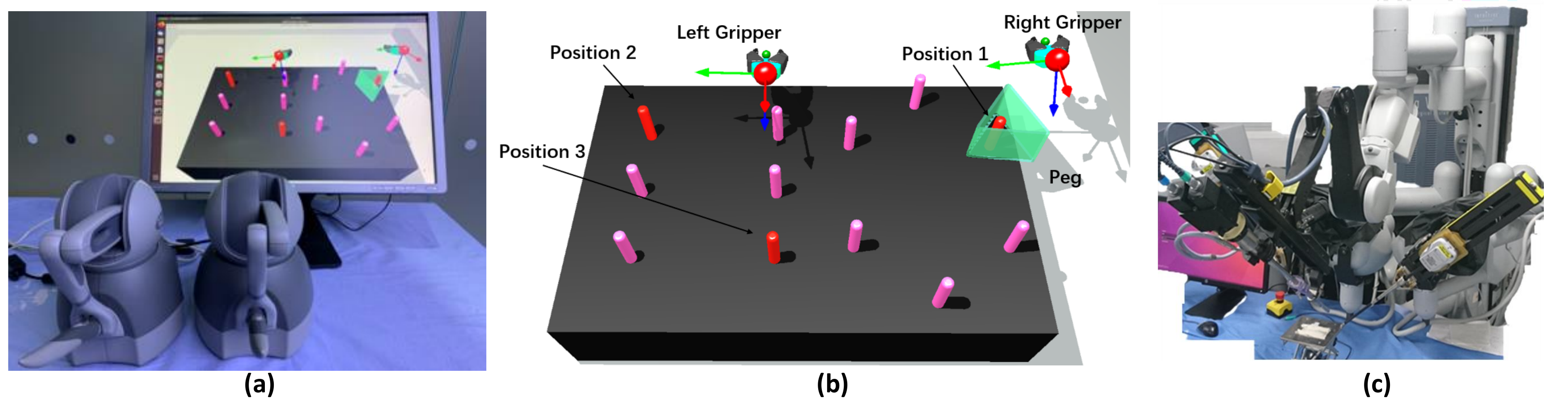} 
	\caption{ Overview of the simulator and physical platform for evaluation in this paper. (a) The experimental setup of the simulator and the two
	Geomagic Touch  motion capture devices  used for teleoperation. (b) The layout of the simulator for peg transfer task and the sketch for the illustration of the experimental protocol. (c) The physical platform used for user studies. } 
	\label{Fig.11} 
\end{figure*}

A simulator targeted for surgical robotic applications has been developed \cite{adnan2019}, and was used  for human demonstration data collection.  Two Geomagic Touch motion capture devices were used as remote controllers for teleoperating the simulated surgical robot, as shown in Fig. \ref{Fig.11} (a).  	Eight participants were recruited in the data collection process after sufficient training, while  36 trials were collected in total.



During data collection, participants were required to  perform the peg transfer task by placing the peg onto the three targeted positions in red subsequently
(see Fig. \ref{Fig.11} (b)).   The initial positions of the grippers were fixed to ensure the comparisons between different trials were fair.  The kinematic data was recorded, including the 6 Degrees-of-Freedom (DoFs) pose sequences of the left and right surgical tool end-effector.
The  procedures for the peg transfer task are as follows:

\begin{itemize}
	\item Control the right surgical tool to grasp the peg from Position 1, transfer it to the left surgical tool and place it on Position 2. 
	
	\item Control the left surgical tool to pick up the peg from Position 2 and place it on Position 3. 
	
	\item Control the right surgical tool to pick up the peg from Position 3,  then locate the peg on Position 1.
\end{itemize}

\subsection{Spatial and Temporal Registration}
\label{registration}
Among the recorded trajectories, the start-point, the end-point of the peg in 3-dimensional (3D) space,  as well as the completion time of all the trials are different. Thus, spatial and temporal registration for these trajectories is necessary. The Iterative Closest Point (ICP) algorithm and Dynamic Time Warping (DTW) algorithm are applied to implement the spatial registration and temporal registration respectively.

\subsubsection{Spatial Registration}


The ICP is normally used to perform geometric registration, especially the alignment of two point-clouds \cite{bouaziz2013sparse}. We use ICP for spatial registration of the recorded trajectories  in this paper.  The initial alignment of two trajectories is implemented by subtracting the  centroid of each trajectory respectively. 
Then the rotation is determined by singular value decomposition. Once the optimal rotation is determined, the translation could be determined as well.
The calculation process will repeat iteratively until the convergency rule is satisfied, which presents the correct alignment of those trajectories. 


\subsubsection{Temporal Registration}

The DTW algorithm is a technique for registering two signal sequences, especially those whose lengths are different from each other \cite{muller2007dynamic}. It calculates the Euclidean distance between sampled points from two sequences as the index of similarity while shortening or extending the sequences in the meantime. With DTW,  optimal matching between two sequences can be found.

\subsection{Gaussian Process Regression}
\label{GPR}
GPR is a probabilistic supervised machine learning method that has been proved to be data-efficient and effective for regression \cite{chen2020supervised}.
GPR can make predictions of desired trajectories given a set of training data and new inputs. Therefore, after spatial and temporal registration, the GPR is employed to generate the desired trajectory based on the registered trajectories.

The prior distribution is assumed to be of Gaussian distribution:
\begin{equation}
\label{GPEquation2}
\begin{aligned}
f(x) \sim GP(m(x),k(x,x'))
\end{aligned}
\end{equation}
where $m(x)=\mathbb{E}[f({x})]$ is a mean function and $k(x,x')=\mathbb{E}\left[(f({x})-m({x}))\left(f\left({x}^{\prime}\right)-m\left({x}^{\prime}\right)\right)\right]$ is a covariance function.
The squared exponential kernel is used for the covariance function. Suppose that $x$ represents the observed data points, while 
$x^{*}$ represents the new inputs.
 The joint distribution of the training outputs $f$ and the predicted outputs $f^*$ can be constructed. 
The predictive distribution can be generated based on conditioning the joint Gaussian prior distribution on the observations \cite{rasmussen2006gaussian}.
Therefore, the predictive equation for GPR can be derived by:
\begin{equation}
\label{GPEquation3}
\begin{aligned}
f^{*} \mid x, f, x^{*} \sim \mathcal{N}\left(\bar{f^{*}}, \Sigma^{*}\right)
\end{aligned}
\end{equation}
where the  expected values are the means $\bar{f^{*}}$, while variances of the predictions can be obtained from the diagonal of the covariance matrix $\Sigma^{*}$. The theory for GPR and the calculation are detailed in \cite{rasmussen2006gaussian}.



\begin{figure}[tb]
	\centering
	\includegraphics[width = 0.95\hsize]{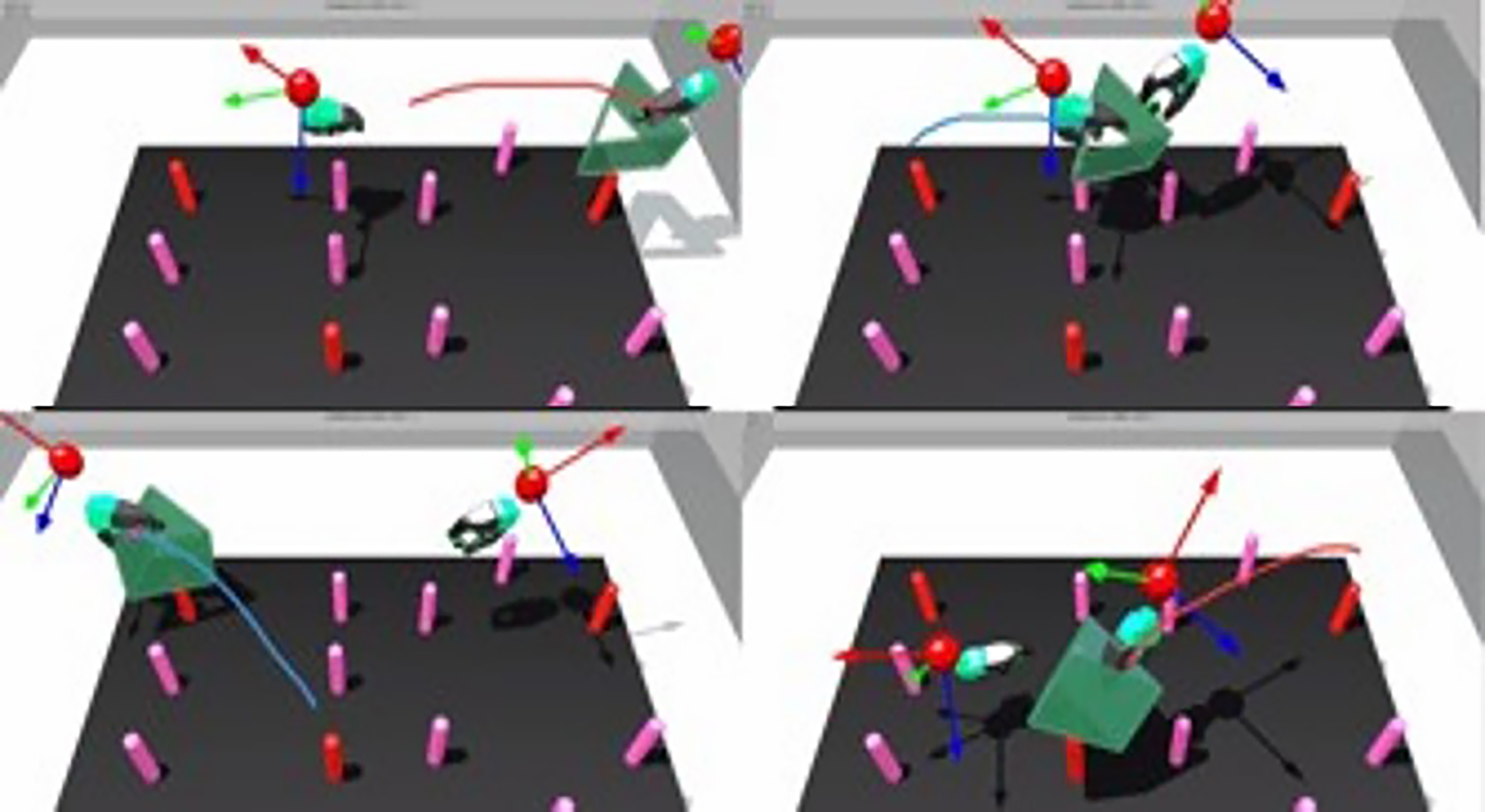}
	\caption{ Visualization of the desired trajectory generated  based on LfD.  The
red curve represents the trajectory of the right hand operation
while the blue curve represents the trajectory of left hand
operation.} 
	\label{fig:GprSim2Real.png}
\end{figure}

For the 3D trajectories, multiple GPRs are applied independently.
To this end, the desired  trajectory of the surgical robot end-effector in 3D space can be generated for autonomous control \cite{chen2020supervised} in some of the surgical sub-tasks, which is essential for the shared control framework.
Fig. \ref{fig:GprSim2Real.png} shows the desired trajectory generated based on LfD in the simulation environment after trajectory registration and regression.

\subsection{Trajectory Adaptation}

To implement the proposed framework on a physical robotic platform via sim-to-real transfer, trajectory adaptation is an essential step.
Since the peg transfer task is goal-driven, the current position of the surgical tool and the target position shall be identified. In this way, the desired trajectory can be adapted to enable the transfer of the peg from the current position to the targets.

 DMP is an effective approach to generalize movements from demonstration. To this end,  DMP is used to enable the robot to perform trajectory adaptation after specifying a new starting point and the targeted position for the peg transfer task.


Let  $\bm{X_{t}}$ denote the position,  $\bm{\dot{X_{t}}}$ denote the  velocity and $\bm{\ddot{X}_{t}}$ denote the acceleration.
$\bm{X_{0}}$ presents the initial position while $\boldsymbol{g}$  represents the goal position. A DMP can be formulated by \eqref{1}.
\begin{equation}
\label{1}
\begin{array}{c}
\gamma \bm{\ddot{X}_{t}}=\alpha_{z}\left(\beta_{z}(\boldsymbol{g}-\bm{X_{t}})-\gamma \bm{\dot{X_{t}}}\right)+\operatorname{diag}\left(\boldsymbol{g}-\bm{X_{0}}\right) \boldsymbol{F}(s_{t}) 
\end{array}
\end{equation}
where $\alpha_{z}, \beta_{z}$ denote scale factors,  $\operatorname{diag}\left(\boldsymbol{g}-\boldsymbol{X}_{0}\right)$ is a diagonal matrix with its elements equal to the coordinates of $\boldsymbol{g-X_{0}}$. ${\gamma}$ is a scalar temporal scaling term, which traditionally is set equal to the duration of the motion. $F(s_{t})$ is a nonlinear function which determines the motion pattern and enables the generation of arbitrary complex movements. 
Variable $s_{t}$ is a monotonically decreasing phase variable (asymptotically decays from 1 to 0), whose dynamics can be known as a first order  canonical system and can be given by $\gamma \dot{s_{t}}=-\alpha_{x} s_{t}$, where $\alpha_{x}$ denotes a scaling factor.

DMP is motivated by the dynamics of a damped spring attached to a goal position with friction. To formulate the DMP in the form of spring-damper system, \eqref{1} can also be expressed as:
\begin{equation}
\bm{\ddot{X}_{t}}=K_{p}\left(\boldsymbol{g}-\bm{X_{t}}\right)-K_{v} \bm{\dot{X}_{t}}+\frac{1}{\gamma}\operatorname{diag}\left(\boldsymbol{g}-\bm{X_{0}}\right)\boldsymbol{F}\left(s_{t}\right)
\end{equation}
where $K_{p}=\frac{1}{\gamma}\alpha_{z}\beta_{z}$ and $K_{v} = \alpha_{z}$ are the stiffness matrix and damping term of DMP in $3 \mathrm{D}$  space respectively. 
 $K_{p}\left(\bm{g}-\bm{X_{t}}\right)-K_{v} \bm{\dot{X}_{t}}$ represents linear spring damper part  while $\boldsymbol{F}\left(s_{t}\right)$ represents the nonlinear part, which can be calculated as follows.

\begin{equation}
\boldsymbol{F}(s_{t})=\frac{\sum_{i}^{N} \boldsymbol{w}_{i} \Psi_{i}(s_{t}) s_{t}}{\sum_{i}^{N} \Psi_{i}(s_{t})}
\end{equation}
with $\boldsymbol{w}_{i}$ represents the weights of the $N$ Gaussian kernel functions $\Psi_{i}(s_{t}):$
\begin{equation}
\Psi_{i}(s_{t})=\exp \left[-h_{i}\left(s_{t}-c_{i}\right)^{2}\right]
\end{equation}
where $h_{i} = -\frac{1}{2 \sigma_{i}}$,  $\sigma_{i}$ is the bandwidth  of the $i_{th}$ Gaussian kernels, while $c_{i}$ represents the kernels' centers distributed to the interval $[0,1]$.

\subsection{Surgical Scene Monitoring}

To enable the transfer of the learned manoeuvres from a simulator to a physical robotic system, trajectory spatial transformation is required, while the start point and the goal point for the desired trajectory should be identified. Hence, a cognitive system should be developed to obtain the relevant information based on the real-time visual information acquired by the stereo laparoscopic system.  

\begin{figure}[tb]
	\centering
	\includegraphics[width = 0.95\hsize]{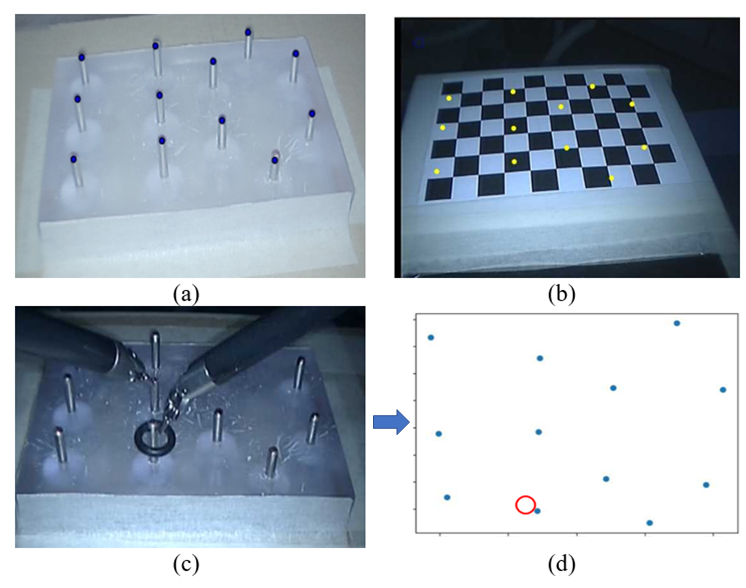}
	\caption{Feature extraction and hand-eye coordination. (a) Features extraction based on ORB and clustering based on GMM. (b) Features mapping onto a referenced chessboard. (c) The experimental scene where the surgical tools were approaching the critical areas for operation. (d) An example showing that the surgical tool reaches the target for local operation.}
	\label{fig:vision}
\end{figure}

The Oriented FAST and rotated BRIEF (ORB) algorithm provided by OpenCV is adopted for feature extraction in this paper \cite{rublee2011orb}. The positions of the feature points can be extracted from the images captured by the endoscope (see the blue dots in Fig. \ref{fig:vision} (a)). 
Gaussian Mixture Model (GMM) is utilized to cluster the scattered points and register them with the chessboard coordinate, as shown in Fig. \ref{fig:vision} (b). The clustering centres represent critical areas for placing the peg. 

Fig. \ref{fig:vision} (c) shows an example of an experimental scene where the surgical tools were approaching the critical areas.  Hand-eye coordination can be used to register the kinematic data of the surgical robot  with stereo images obtained by the vision system. The extrinsic parameters of the camera with respect to the object can be obtained through the chessboard calibration. The goal position obtained from the image frame $p_t$ can be converted to the world coordinate. Therefore,  the start point  $\bm{X_0}$ and goal position $\boldsymbol{g}$ can be obtained, while the trajectory adaptation based on DMP can be achieved. Fig. \ref{fig:vision} (d) demonstrates an example when the surgical tool reaches the goal for local operation. The red circle indicates the position of the end-effector of the surgical tool.

\subsection{ Context-Awareness for Role Adaptation}
\label{context}


In this paper, context-awareness of the surgical scene is developed based on a convolutional neural network (CNN) to allow the relative roles of the robot and the human operator to be adjusted during the surgical operation.
It can be used to determine the switching conditions among different operation phases, which depends on the characteristics of the corresponding recognized contexts. 
Suppose that $F(.)$ is the neural network used for context-awareness, the probability of the output of $F(.)$  is $\mathcal{P}(c = c^{*})(c^{*} = 0,1,2)$. The surgical scene can be classified to three contexts, including ``move to next target" $(c = 0)$, ``bimanual operation" $(c = 1)$ and ``local operation" $(c = 2)$. The constructed database was manually annotated with the context labels defined at the frame level for model training. 70\% of data was used for training while 30\% of data was used for testing.

During the image preprocessing procedure, the RGB images were converted to grayscale images and were resized to 150 ${\times}$ 150. The network consists of 6 convolutional layers and 2 fully-connect layers.  The categorical cross-entropy was used as our objective cost function for training the network, while Adam optimizer was used  with a learning rate of 0.0001 \cite{kingma2014adam}.    An early stopping strategy was implemented for  model training.

\begin{figure*}[tb]
	\centering
	\includegraphics[width = 1\hsize]{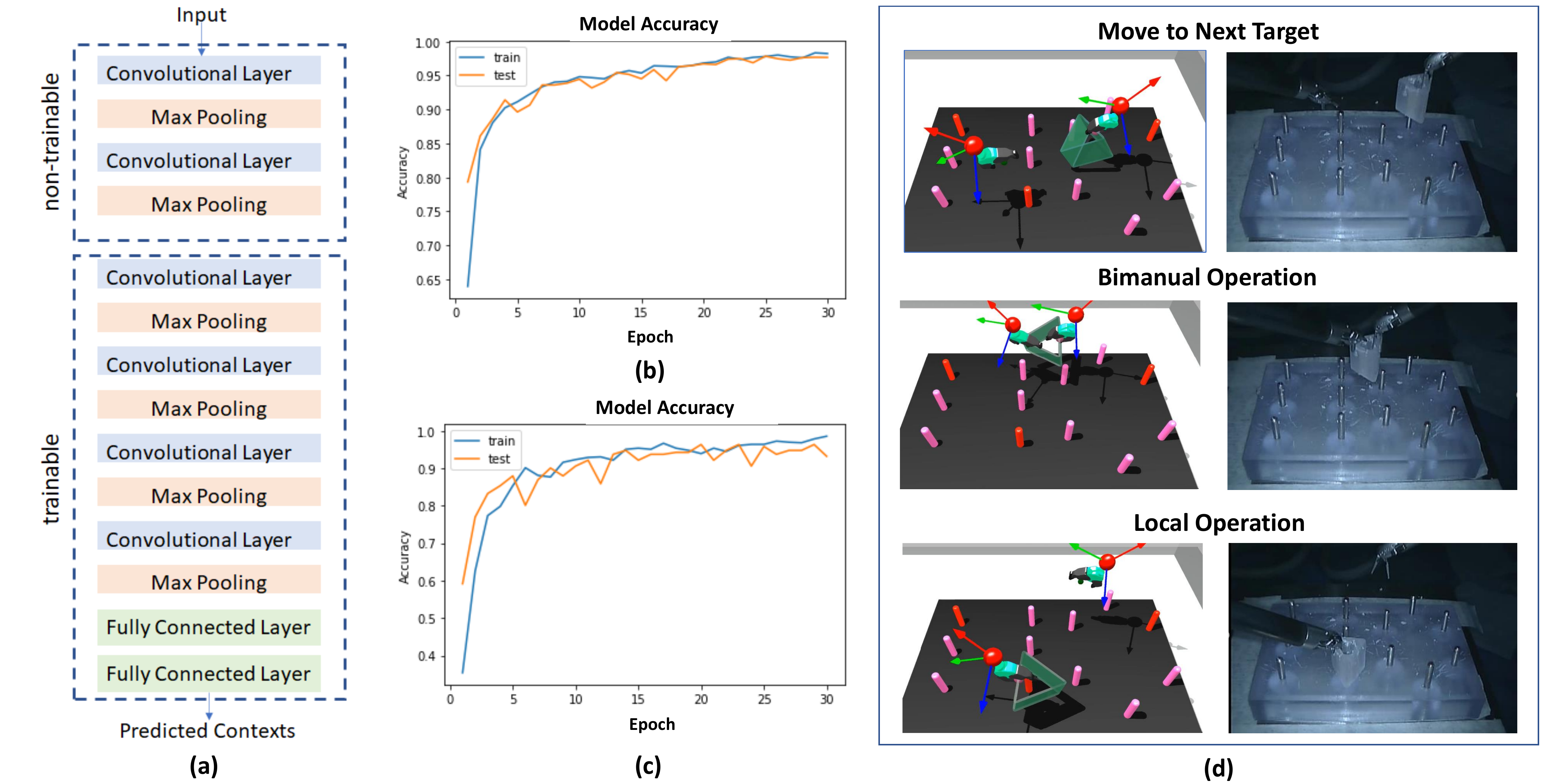}
	\caption{The model construction, the overall accuracy and the experimental results for context-awareness. (a) The neural network architecture for context-awareness. (b) and (c) show the model training and testing accuracy for context-awareness when using the simulation data and experimental data respectively. (d) Examples of  three operation scenes with different contexts for the  peg transfer task in simulation and physical environment respectively.}
	\label{fig:transfer results1}
\end{figure*}



 Fig. \ref{fig:transfer results1} (a) shows the architecture of the CNN model for context-awareness. The overall results for the model accuracy during the training and testing process for scenes recognition are shown in Fig.~\ref{fig:transfer results1} (b). After training 20 epochs, the testing accuracy for scene recognition can reach 0.977.  
 For online deployment, transfer learning can be used to fine-tune the learned model obtained in the simulator and applied to the physical robotic system after calibration \cite{zhang2020automatic}. More details will be illustrated in Section \ref{results-sim2real}.


\subsection{Model Execution}

After model training of $F(.)$, the human-robot shared control framework can then be implemented by a weighted summation of the commands generated by both the human operator and the intelligent surgical robot enabled by LfD.

Suppose that $\Delta{P^s_t}$ is the position increment value for the surgical robot control; $\Delta{P^r_t}$ is the remote controller's end-effector position increment determined by the LfD model; $\Delta{P^h_t}$ represents the control command generated by human operators via remote controllers. 
Let $\tau$ denote the motion scaling factor \cite{zhang2018self,zhang2020microsurgical}.  $\alpha{\in}[0,1]$ is a weight parameter regulated by the real-time context-awareness of the surgical operation scenes.  The position control can be determined by \eqref{eq:1}, while the orientation of the surgical robot's end-effectors is set to be the same as the remote controllers' end-effectors to ensure intuitive control \cite{Zhang2020Hamlyn,zhang2020ergonomic}.
\begin{gather}
\label{eq:1}
\Delta{P^s_t} = {\tau}[\alpha \Delta{P^h_t} + (1-\alpha)\Delta{P^r_t}  ]
\end{gather}

For applications on physical robotic platforms, to achieve a smooth transition between different contexts, the surgical operation status is expressed in the form of probability. At time step $t$, $c(t) = \textit{argmax}\{\mathcal{P}(c=c^{*})(c^{*}=0,1,2)\}$, when $\mathcal{P}(c={c}(t))$  is significantly greater than $\mathcal{P}(c=c({t-1}))$($c(t){\neq}c({t-1})$), we assume that the operation scene is switched to another. That is to say, when $\mathcal{P}(c={c}(t))$ is sufficiently leading, the estimation of ${c}(t)$ has been stable and can be well trusted.  Otherwise, the operation may be at the transition process between different contexts. 
Let $\lambda$ denote the threshold value, ${\Delta}\mathcal{P}_c={\mathcal{P}(c={c}(t)) - \mathcal{P}(c=c({t-1}))}$. 	Hence, $\alpha$ can be determined by \eqref{alpha}.
	\begin{equation}
	\label{alpha}
	\alpha = \left\{
		\begin{array}{ll}
	\mathcal{P}(c=c({t-1})) & \mathrm{if}~{\Delta}\mathcal{P}_c < \lambda, {c}(t)=1 or 2 \\
	1 -	\mathcal{P}(c=c({t-1})) & \mathrm{if}~{\Delta}\mathcal{P}_c < \lambda, {c}(t)=0 \\
			1 & \mathrm{if}~{\Delta}\mathcal{P}_c \geq \lambda, {c}(t)=1 or 2 \\
			0 & \mathrm{if}~{\Delta}\mathcal{P}_c \geq \lambda, {c}(t)=0
		\end{array} \right. 
	\end{equation}
	where  ${\lambda}=0.5$ is used in this paper.  ${\lambda} \in (0,1)$ can be tuned when it is applied to different surgical tasks.

The role adaptation mechanism can enable the seamless switching among three control modes, i.e. manual control, autonomous control and adaptive shared control. If the operation scene is recognized as ``bimanual operation"  $(c(t) = 1)$ or ``local operation" $(c(t) = 2)$, $\alpha$ is set to be $1$. This means that the manual control is required for fine motion generation, since the inaccuracy caused by the machine learning model should be avoided. If the context is recognized as ``moving to the next target"  $(c(t) = 0)$,  $\alpha$ is set to be $0$. This means that during relocation of the surgical tools, we assume that the operation does not need human engagement, since the required precision degree for the control in this context is not high. In this case, the autonomous control mode for some of the surgical sub-tasks can be achieved. If the operation is at the transition process between different contexts, the adaptive shared control mode will be implemented with $\alpha{\in}(0,1)$.

\section{User Studies}
\label{UserStudies}

\subsection{Experimental Setup}

In order to verify the effectiveness of the proposed framework on a physical robot, user studies were conducted on the da Vinci Research Kit (dVRK) \cite{chen2017software} based on the peg transfer task, as shown in Fig. \ref{Fig.11} (c).  
During the user studies, the participants used two Geomagic Touch motion capture devices  to control the da Vinci Robot. The experiments on the dVRK used a peg board with the same dimension as the one in the simulator, while the experimental protocols remained the same.

Eight subjects were invited to join the user studies. Six of the subjects have teleoperation experience.   All of them had a practice session to get familiar with the experimental protocols before conducting the experiments for three to five trials.  The kinematics data of the surgical robot was recorded during the user studies, and can be used to compute the evaluation metrics.

\subsection{Online Deployment of the Proposed Framework}
\label{results-sim2real}

During the transfer learning process, the parameters of the first two convolutional layers were fixed. This means that we don't need to train a completely new model from scratch. We assume that the feature extraction mode can be similar. The fixed layers and the trainable layers are illustrated in  Fig.~\ref{fig:transfer results1} (a).
The model training accuracy for transfer learning is plotted in Fig.~\ref{fig:transfer results1} (c), where the testing accuracy is 0.932  when using transfer learning.  Fig.~\ref{fig:transfer results1} (d) shows the examples of three operation scenes for the peg transfer task, which were collected from the simulator and the dVRK for the evaluation of the accuracy of context-awareness in both simulator and physical system.


For deployment on a physical robotic system, DMP was used for sim-to-real trajectory adaptation. Following that, with real-time accurate  predictions of the contexts for surgical operation scenes, the online automatic switching between different operation phases with different levels of human engagement during the shared control process can therefore be achieved. 
To this end, the adaptive human-robot shared control framework can  be constructed for user studies.


\subsection{Results Analysis for dVRK Based User Studies}

The path length of the remote controller ($M (m)$), task completion time ($T (s)$), the average velocity of the surgical robot ($A (mm/s)$) and the total clutching number ($C$) are used as the evaluation metrics.  Normality tests (Shapiro-Wilk test) were conducted to identify whether the evaluation metrics have non-parametric nature or not, where 0.05 is set as the significance level. Since $A (mm/s)$ and $C$ satisfy the normal distribution assumption, T-tests were conducted
to determine if there is a significant difference between the means of the manual control results and the shared control results.
Wilcoxon signed-rank tests were conducted for non-parametric statistical comparison between $M (m)$ and $T (s)$. A p-value$<$0.05 is considered significant.

\begin{figure}[tb]
	\centering
	\includegraphics[width = 0.95\hsize]{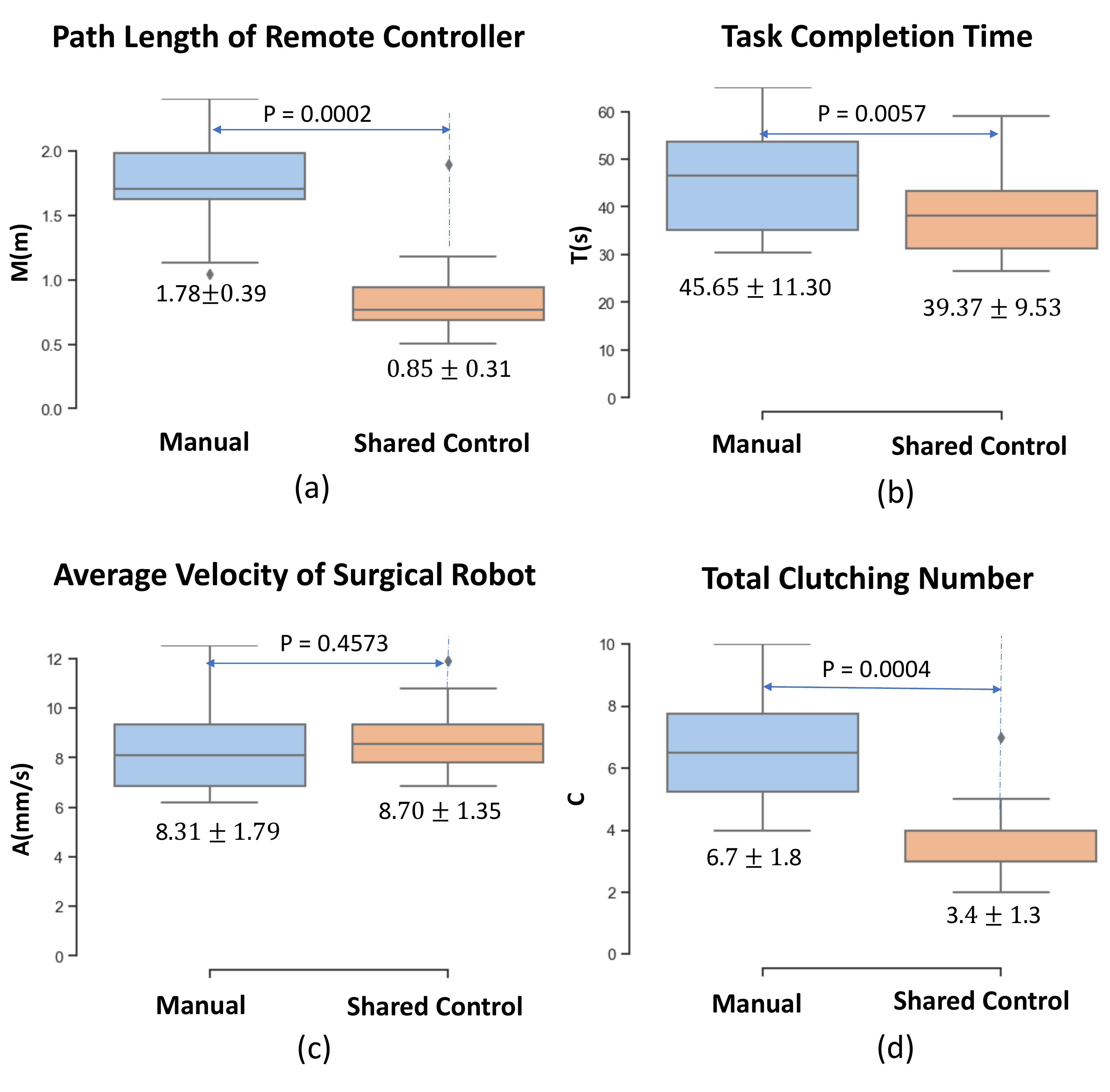}
	\caption{The box plot results for the  user studies based on the dVRK platform. Comparisons in terms of (a) path length of the remote controller, (b) task completion time, (c)  average velocity of surgical robot, (d) total clutching number.} 
	\label{fig:box1dvrk}
\end{figure}

The comparative results for the human-robot shared control and the manual control via teleoperation based on the dVRK system are shown in Fig. \ref{fig:box1dvrk}. With the human-robot shared control, the path length of the remote controller and the total clutching number can be reduced significantly by about  50\%. The average task completion time is reduced from 45.7$s$ to 39.3$s$. The differences between the path length of the remote controller, the total clutching number and the task completion time are significant ($p<0.05$). The average velocity of the surgical robot is higher (human-robot shared control vs. manual control via teleoperation), but the difference is not significant.

\section{Conclusions}\label{Conclusions}

In this paper, surgical robots can learn to perform the surgical operation with human in a shared control manner. The desired trajectory was learned via  GPR after spatial and temporal registration of the demonstration data.  
DMP was used  for sim-to-real trajectory adaptation. 
With this approach, most of the training processes can be conducted based on the simulator and then the desired trajectory obtained via GPR can be transferred to a physical robotic system to support the robot to generate commands automatically during task execution.  For online deployment, context-awareness is used to realize role adaptation between human and robot, which leads to adaptive human-robot shared control. Human operator commands are generated via remote controllers, while robot commands are determined by the desired trajectory obtained via LfD with sim-to-real adaptation. By incorporating the context information, the weight parameter can be adjusted to combine human operator commands and robot commands in an optimized manner.  The  knowledge extracted from the simulator for context-awareness can be transferred to physical system after fine-tuning the neural network model.  The human-robot shared control framework was evaluated on the dVRK with user studies. The experimental results demonstrated that the performances of the subjects using shared control mode outperformed those using traditional manual control mode.

Future work will include applying this technique to more complex surgical operations, such as suturing and knot tying.  The generalizability of the proposed framework will be further verified in scenarios with consideration of obstacles avoidance.
In the meantime, haptic guidance will be incorporated into the control framework. More advanced machine learning methods can be explored for LfD and sim-to-real adaptation.

\bibliographystyle{IEEEtran}
\bibliography{references}

\begin{thebibliography}{10}
\providecommand{\url}[1]{#1}
\csname url@samestyle\endcsname
\providecommand{\newblock}{\relax}
\providecommand{\bibinfo}[2]{#2}
\providecommand{\BIBentrySTDinterwordspacing}{\spaceskip=0pt\relax}
\providecommand{\BIBentryALTinterwordstretchfactor}{4}
\providecommand{\BIBentryALTinterwordspacing}{\spaceskip=\fontdimen2\font plus
\BIBentryALTinterwordstretchfactor\fontdimen3\font minus
  \fontdimen4\font\relax}
\providecommand{\BIBforeignlanguage}[2]{{%
\expandafter\ifx\csname l@#1\endcsname\relax
\typeout{** WARNING: IEEEtran.bst: No hyphenation pattern has been}%
\typeout{** loaded for the language `#1'. Using the pattern for}%
\typeout{** the default language instead.}%
\else
\language=\csname l@#1\endcsname
\fi
#2}}
\providecommand{\BIBdecl}{\relax}
\BIBdecl

\bibitem{yang2018grand}
G.-Z. Yang, J.~Bellingham, P.~E. Dupont, P.~Fischer, L.~Floridi, R.~Full,
  N.~Jacobstein, V.~Kumar, M.~McNutt, R.~Merrifield \emph{et~al.}, ``The grand
  challenges of science robotics,'' \emph{Science robotics}, vol.~3, no.~14, p.
  eaar7650, 2018.

\bibitem{gao2021progress}
A.~Gao, R.~R. Murphy, W.~Chen, G.~Dagnino, P.~Fischer, M.~G. Gutierrez,
  D.~Kundrat, B.~J. Nelson, N.~Shamsudhin, H.~Su \emph{et~al.}, ``Progress in
  robotics for combating infectious diseases,'' \emph{Science Robotics},
  vol.~6, no.~52, p. eabf1462, 2021.

\bibitem{zhang2019design}
D.~Zhang, J.~Liu, L.~Zhang, and G.-Z. Yang, ``Design and verification of a
  portable master manipulator based on an effective workspace analysis
  framework,'' in \emph{2019 IEEE/RSJ International Conference on Intelligent
  Robots and Systems (IROS)}.\hskip 1em plus 0.5em minus 0.4em\relax IEEE,
  2019, pp. 417--424.

\bibitem{Zhang2020Hamlyn}
D.~Zhang, J.~Liu, L.~Zhang, and G.~Z. Yang, ``Hamlyn crm: a compact master
  manipulator for surgical robot remote control,'' \emph{International Journal
  of Computer Assisted Radiology \& Surgery}, vol.~15, no.~3, pp. 503--514,
  2020.

\bibitem{zhang2019handheld}
D.~Zhang, Y.~Guo, J.~Chen, J.~Liu, and G.-Z. Yang, ``A handheld master
  controller for robot-assisted microsurgery,'' in \emph{2019 IEEE/RSJ
  International Conference on Intelligent Robots and Systems (IROS)}.\hskip 1em
  plus 0.5em minus 0.4em\relax IEEE, 2019, pp. 394--400.

\bibitem{peters2018review}
B.~S. Peters, P.~R. Armijo, C.~Krause, S.~A. Choudhury, and D.~Oleynikov,
  ``Review of emerging surgical robotic technology,'' \emph{Surgical
  endoscopy}, vol.~32, no.~4, pp. 1636--1655, 2018.

\bibitem{yang2017medical}
G.-Z. Yang, J.~Cambias, K.~Cleary, E.~Daimler, J.~Drake, P.~E. Dupont, N.~Hata,
  P.~Kazanzides, S.~Martel, R.~V. Patel \emph{et~al.}, ``Medical
  robotics—regulatory, ethical, and legal considerations for increasing
  levels of autonomy,'' \emph{Sci. Robot}, vol.~2, no.~4, p. 8638, 2017.

\bibitem{payne2021shared}
C.~J. Payne, K.~Vyas, D.~Bautista-Salinas, D.~Zhang, H.~J. Marcus, and G.-Z.
  Yang, ``Shared-control robots,'' \emph{Neurosurgical Robotics}, pp. 63--79,
  2021.

\bibitem{li2013building}
Y.~Li, K.~P. Tee, S.~S. Ge, and H.~Li, ``Building companionship through
  human-robot collaboration,'' in \emph{International Conference on Social
  Robotics}.\hskip 1em plus 0.5em minus 0.4em\relax Springer, 2013, pp. 1--7.

\bibitem{amirshirzad2019human}
N.~Amirshirzad, A.~Kumru, and E.~Oztop, ``Human adaptation to human--robot
  shared control,'' \emph{IEEE Transactions on Human-Machine Systems}, vol.~49,
  no.~2, pp. 126--136, 2019.

\bibitem{zhang2021explainable}
D.~Zhang, Q.~Li, Y.~Zheng, L.~Wei, D.~Zhang, and Z.~Zhang, ``Explainable
  hierarchical imitation learning for robotic drink pouring,'' \emph{IEEE
  Transactions on Automation Science and Engineering}, 2021.

\bibitem{chen2020supervised}
J.~Chen, D.~Zhang, A.~Munawar, R.~Zhu, B.~Lo, G.~S. Fischer, and G.-Z. Yang,
  ``Supervised semi-autonomous control for surgical robot based on bayesian
  optimization,'' in \emph{2020 IEEE/RSJ International Conference on
  Intelligent Robots and Systems (IROS)}.\hskip 1em plus 0.5em minus
  0.4em\relax IEEE, pp. 2943--2949.

\bibitem{ravichandar2020recent}
H.~Ravichandar, A.~S. Polydoros, S.~Chernova, and A.~Billard, ``Recent advances
  in robot learning from demonstration,'' \emph{Annual Review of Control,
  Robotics, and Autonomous Systems}, vol.~3, pp. 297--330, 2020.

\bibitem{power2015cooperative}
M.~Power, H.~Rafii-Tari, C.~Bergeles, V.~Vitiello, and G.-Z. Yang, ``A
  cooperative control framework for haptic guidance of bimanual surgical tasks
  based on learning from demonstration,'' in \emph{2015 IEEE International
  Conference on Robotics and Automation (ICRA)}.\hskip 1em plus 0.5em minus
  0.4em\relax IEEE, 2015, pp. 5330--5337.

\bibitem{shin2019autonomous}
C.~Shin, P.~W. Ferguson, S.~A. Pedram, J.~Ma, E.~P. Dutson, and J.~Rosen,
  ``Autonomous tissue manipulation via surgical robot using learning based
  model predictive control,'' in \emph{2019 International Conference on
  Robotics and Automation (ICRA)}.\hskip 1em plus 0.5em minus 0.4em\relax IEEE,
  2019, pp. 3875--3881.

\bibitem{van2010superhuman}
J.~Van Den~Berg, S.~Miller, D.~Duckworth, H.~Hu, A.~Wan, X.-Y. Fu, K.~Goldberg,
  and P.~Abbeel, ``Superhuman performance of surgical tasks by robots using
  iterative learning from human-guided demonstrations,'' in \emph{2010 IEEE
  International Conference on Robotics and Automation}.\hskip 1em plus 0.5em
  minus 0.4em\relax IEEE, 2010, pp. 2074--2081.

\bibitem{schulman2013case}
J.~Schulman, A.~Gupta, S.~Venkatesan, M.~Tayson-Frederick, and P.~Abbeel, ``A
  case study of trajectory transfer through non-rigid registration for a
  simplified suturing scenario,'' in \emph{2013 IEEE/RSJ International
  Conference on Intelligent Robots and Systems}.\hskip 1em plus 0.5em minus
  0.4em\relax IEEE, 2013, pp. 4111--4117.

\bibitem{li2020skill}
J.~Li, Z.~Li, X.~Li, Y.~Feng, Y.~Hu, and B.~Xu, ``Skill learning strategy based
  on dynamic motion primitives for human--robot cooperative manipulation,''
  \emph{IEEE Transactions on Cognitive and Developmental Systems}, vol.~13,
  no.~1, pp. 105--117, 2020.

\bibitem{wang2021real}
R.~Wang, D.~Zhang, Q.~Li, X.-Y. Zhou, and B.~Lo, ``Real-time surgical
  environment enhancement for robot-assisted minimally invasive surgery based
  on super-resolution,'' in \emph{2021 IEEE International Conference on
  Robotics and Automation (ICRA)}.\hskip 1em plus 0.5em minus 0.4em\relax IEEE,
  2021, pp. 3434--3440.

\bibitem{zhang2021surgical}
D.~Zhang, R.~Wang, and B.~Lo, ``Surgical gesture recognition based on
  bidirectional multi-layer independently rnn with explainable spatial feature
  extraction,'' in \emph{2021 IEEE International Conference on Robotics and
  Automation (ICRA)}.\hskip 1em plus 0.5em minus 0.4em\relax IEEE, 2021, pp.
  1350--1356.

\bibitem{adnan2019}
A.~{Munawar}, Y.~{Wang}, R.~{Gondokaryono}, and G.~S. {Fischer}, ``A real-time
  dynamic simulator and an associated front-end representation format for
  simulating complex robots and environments,'' in \emph{2019 IEEE/RSJ
  International Conference on Intelligent Robots and Systems (IROS)}, Nov 2019,
  pp. 1875--1882.

\bibitem{bouaziz2013sparse}
S.~Bouaziz, A.~Tagliasacchi, and M.~Pauly, ``Sparse iterative closest point,''
  in \emph{Computer graphics forum}, vol.~32, no.~5.\hskip 1em plus 0.5em minus
  0.4em\relax Wiley Online Library, 2013, pp. 113--123.

\bibitem{muller2007dynamic}
M.~M{\"u}ller, ``Dynamic time warping,'' \emph{Information retrieval for music
  and motion}, pp. 69--84, 2007.

\bibitem{rasmussen2006gaussian}
C.~Rasmussen and C.~Williams, ``Gaussian processes for machine learning, ser.
  adaptive computation and machine learning, t. dietterich, ed,'' 2006.

\bibitem{rublee2011orb}
E.~Rublee, V.~Rabaud, K.~Konolige, and G.~Bradski, ``Orb: An efficient
  alternative to sift or surf,'' in \emph{2011 International conference on
  computer vision}.\hskip 1em plus 0.5em minus 0.4em\relax Ieee, 2011, pp.
  2564--2571.

\bibitem{kingma2014adam}
D.~P. Kingma and J.~Ba, ``Adam: A method for stochastic optimization,''
  \emph{arXiv preprint arXiv:1412.6980}, 2014.

\bibitem{zhang2020automatic}
D.~Zhang, Z.~Wu, J.~Chen, A.~Gao, X.~Chen, P.~Li, Z.~Wang, G.~Yang, B.~Lo, and
  G.-Z. Yang, ``Automatic microsurgical skill assessment based on cross-domain
  transfer learning,'' \emph{IEEE Robotics and Automation Letters}, vol.~5,
  no.~3, pp. 4148--4155, 2020.

\bibitem{zhang2018self}
D.~Zhang, B.~Xiao, B.~Huang, L.~Zhang, J.~Liu, and G.-Z. Yang, ``A
  self-adaptive motion scaling framework for surgical robot remote control,''
  \emph{IEEE Robotics and Automation Letters}, vol.~4, no.~2, pp. 359--366,
  2018.

\bibitem{zhang2020microsurgical}
D.~Zhang, J.~Chen, W.~Li, D.~B. Salinas, and G.-Z. Yang, ``A microsurgical
  robot research platform for robot-assisted microsurgery research and
  training,'' \emph{International journal of computer assisted radiology and
  surgery}, vol.~15, no.~1, pp. 15--25, 2020.

\bibitem{zhang2020ergonomic}
D.~Zhang, J.~Liu, A.~Gao, and G.-Z. Yang, ``An ergonomic shared workspace
  analysis framework for the optimal placement of a compact master control
  console,'' \emph{IEEE Robotics and Automation Letters}, vol.~5, no.~2, pp.
  2995--3002, 2020.

\bibitem{chen2017software}
Z.~Chen, A.~Deguet, R.~H. Taylor, and P.~Kazanzides, ``Software architecture of
  the da vinci research kit,'' in \emph{2017 First IEEE International
  Conference on Robotic Computing (IRC)}.\hskip 1em plus 0.5em minus
  0.4em\relax IEEE, 2017, pp. 180--187.

\end{thebibliography}

\end{document}